\documentclass[runningheads]{llncs}
\usepackage{graphicx}
\usepackage{mathtools}
\usepackage{times}
\usepackage{latexsym}
\usepackage{textcomp}
\usepackage{xcolor}
\usepackage{array,multirow}
\usepackage{float}
\usepackage{mathtools}
\usepackage{dirtytalk}
\usepackage{times}
\usepackage{latexsym}
\usepackage{breqn}
\usepackage{amssymb,amsfonts}
\usepackage{booktabs} 
\usepackage{amsmath}
\usepackage{bbm}
\usepackage{amssymb,amsfonts}
\usepackage[noend]{algpseudocode}
\usepackage{tabularx}
\usepackage[ruled,vlined]{algorithm2e}
\usepackage{lipsum}
\usepackage{algpseudocode}
\begin{document}
\title{Am I Rare?\\ An Intelligent Summarization Approach for Identifying Hidden Anomalies}
\titlerunning{INSIDENT}
\author{Samira Ghodratnama\inst{1} \and
Mehrdad Zakershahrak\inst{2}\and
Fariborz Sobhanmanesh\inst{1}}

\def\algbackskip{\hskip-\ALG@thistlm}
\makeatother

\authorrunning{S. Ghodratnama et al.}
\institute{Macquarie University, Sydney, Australia
\email{\{samira.ghodratnama,fariborz.sobhanmanesh\}@mq.edu.au} \and
Arizona State University, Arizona, United States\\
\email{\{mehrdad\}@asu.edu}}

\maketitle      
\begin{abstract}
Monitoring network traffic data to detect any hidden patterns of anomalies is a challenging and time-consuming task which requires high computing resources.
To this end, an appropriate summarization technique is of great importance, where it can be a substitute for the original data.
However, the summarized data is under the threat of removing anomalies.
Therefore, it is vital to create a summary that can reflect the same pattern as the original data.
Therefore, in this paper, we propose an INtelligent Summarization approach for IDENTifying hidden anomalies, called \textit{INSIDENT}.
The proposed approach guarantees to keep the original data distribution in summarized data. 
Our approach is a clustering-based algorithm that dynamically maps original feature space to a new feature space by locally weighting features in each cluster.
Therefore, in new feature space, similar samples are closer, and consequently, outliers are more detectable.
Besides,  selecting representatives based on cluster size keeps the same distribution as the original data in summarized data.
\textit{INSIDENT} can be used both as the preprocess approach before performing anomaly detection algorithms and anomaly detection algorithm.
The experimental results on benchmark datasets prove a summary of the data can be a substitute for original data in the anomaly detection task.
\end{abstract}
\section{Introduction}
Monitoring the fast and large volume of Internet traffic data that is being generated is paramount since they may have instances of anomalous network traffic, which makes the system vulnerable.
However, detecting anomalies when we face big data is computationally expensive and still an open challenge.
To this end, summarization is a practical approach that produces a condensed version of the original data.
Therefore, a summary of the network traffic data helps network managers quickly assess what is happening in the network.
For instance,  the summary should still give insight into most visited websites, frequently used applications, and incoming traffic patterns.
In~\cite{hoplaros2014data} authors defined three scenarios in which summarization can help in traffic data, including:
Summarizing network traffic can give an overview of what is going on in the network to the administrator.
Summarized network traffic can be used as input to anomaly detection algorithms to reduce the cost.
A summary of intrusion detection alarms facilitates the administrator's duty.
In all mentioned scenarios, a concise representation of the data helps both the administrator and the analysis algorithms.

Different data summarization techniques are designed for other applications such as transactional data or stream data~\cite{ahmed2019data}, which can be applied to traffic data.
However, they have some drawbacks to be used for anomaly detection purposes, including:
\begin{itemize}
    \item Clustering is the most used approach for summarization, where centers are considered as the summarized data.
 The problem is that the centroids may not be a part of the original data.
    \item Detecting frequent itemsets is another approach which only captures frequent items in the summaries.
    Therefore, they ignore or leave out anomalies that may be infrequent.
    Consequently, anomaly detection techniques do not perform well on summaries as they do not contain any anomalies.
    \item Semantics-based techniques do not keep the same samples in the summarized data. 
    \item Statistical based techniques such as sampling do not guarantee the representation of anomalies in summary since they use a sampling-based summarization technique. 
\end{itemize}
Therefore, not all summarization approaches are proper for anomaly detection purposes.
Consequently, there is a need for an efficient network traffic summarization technique so that the summary more closely resembles the original network traffic.
In this context, summarization aims to create a summary from original data that includes interesting patterns, especially anomalies, and normal data for further analysis.

This paper proposes an intelligent summarization approach suitable for anomaly detection on network traffic datasets, which guarantees the preservation of original data distribution.
We investigate the adaptation of clustering and KNN algorithms to create a summary.
The proposed algorithm is used in two scenarios: i) as the preprocess approach for performing anomaly detection, ii) to detect anomalies in supervised problems as it reveals the hidden structure of data.
The proposed summarization technique can also be adapted to other domains where big data requires being minded for interesting and relevant information.
The rest of this paper is organized as follows. 
Section~\ref{related} discusses the state-of-the-art methods.
Section~\ref{proposed} presents the proposed method, and section~\ref{evaluation} explains the experimental results and justifies the obtained results. 
Finally, section~\ref{conclusion} concludes the paper and discusses future work.
\section{Related Work}
\label{related}
Summarization has been widely explored in many domains and applications, using a variety of techniques~\cite{beheshti2018iprocess,schiliro2018icop,amouzgar2018isheets,ghodratnama2021intelligent,summary2vec,zakershahrak2020we}.
When data size increases, the anomaly detection techniques perform poorly due to increasing false alarms and computational cost.
Detecting anomalies from a summary could address these issues.
However, existing summarization techniques cannot accurately represent the rare anomalies present in the dataset. 
In this section, we will present related work on traffic data summarization, along with anomaly detection techniques.
It is worth mentioning although the general goal is to represent an input dataset in a condensed version, there is no definition of a good summary since each application requires a unique technique.
For anomaly detection purposes, a good summary should be representative of all samples in the original dataset.
\subsection{Network Analysis Tools}
Different network analysis tools summarize network traffic data, such as Traffic Flow Analysis Tool, Flow-tools, Network Visualization Tools, and Network Monitoring Tools~\cite{ahmed2019intelligent}.
They produce a graphical report using different measurements, such as network bandwidth or latency.
However, they only characterize and aggregate traffic instances based on a single attribute, such as the source/destination address or protocol. 
As a result, they are suitable to extract insights, not for further processing tasks such as anomaly detection.
Besides, the objective of a summary is to provide an accurate report of the network's traffic patterns.
Consequently, the summarization technique should identify traffic patterns based on arbitrary combinations of attributes efficiently. 
\subsection{Statistical Approaches}
Statistical approaches aim to estimate the statistical distribution of data that could approximate the data set pattern.
Sampling is a common technique in this category where a  sample is a subset of the dataset.
There are different kinds of sampling in practice, including i) simple random sampling, ii) stratified random sampling, iii) systematic sampling, iv) cluster random sampling, and v) multi-stage random sampling~\cite{cochran1977sampling,ghodratnama2009innovative}.
However, summarized data using sampling is under the threat of removing anomalies.
To solve this problem, in a recent work~\cite{ahmed2019intelligent}, the author proposed a sampling-based summarization technique, called SUCh, which integrated the concept of sampling using the modified Chernoff bound to include anomalous instances in summary.
\textit{SUCh} is computationally effective than the existing techniques and also performs better in identifying rare anomalies.
However, an essential aspect of the summarization is representing all different types of traffic behavior.
Although \textit{SUCh}  ensures the presence of anomalies, it ignores other types of traffic as they focus only on anomalous data.
\subsection{Machine Learning Approaches}
Supervised and unsupervised learning techniques are two widely used knowledge discovery techniques.
Two common machine learning algorithms used in summarizing network traffic data are \textit{frequent itemsets} and \textit{clustering}.
Frequent itemsets are a set of items that appears more frequently than the rest of the samples.
Different algorithms are used to detect frequent itemsets~\cite{chandola2007summarization}.
However, they are proper for detecting frequent items, not rare anomalies.
Two main clustering-based algorithms for network traffic data summarization include centroid-based and feature-wise intersectin clustering algorithms.
In a centroid-based summarization, after clustering samples, centroids are used to form the summary.
Different variations of the k-means algorithm are widely used due to its simplicity, which can handle high-dimensional data~\cite{ghodratnama2015efficient,wendel2005scalable}.
In a feature-wise intersection-based summarization, the summary is created from each cluster using the feature-wise intersection of the data instances after clustering ~\cite{chandola2007summarization,hoplaros2014data}.
Consequently, summaries from all the clusters are combined to produce the final summary.
This approach is best fitted for datasets with identical attribute values and, therefore, not suitable for detecting rare anomalies.
\subsection{Semantic-based Approaches}
Semantic-based approaches are not suitable for anomaly detection since they do not produce a summary, which is part of the original data.
Examples are linguistic summaries, which are based on the fuzzy.
These approaches produce natural language expressions that describe important facts about the given data to enhance the human understanding of the network traffic summaries~\cite{pouzols2011summarization}.
Attribute Oriented Induction (AOI) is another semantic-based approach
aims to describe data in a concise and general manner~\cite{han199616}. 
AOI is a generalization process that abstracts a large dataset from a low conceptual level to a relatively higher conceptual level. 
Other semantic-based approaches include Fascicles~\cite{jagadish1999semantic}, which relies on an extended form of association rules and perform lossy semantic compression.
SPARTAN is another semantic-based summarization technique~\cite{babu2001spartan}, which generalizes the fascicles approach. 
\subsection{Anomaly Detection Techniques}
Anomaly detection is an important data analysis task that detects anomalous or abnormal data from a given dataset.
Anomalies are patterns in data that do not follow the well-defined characteristic of typical patterns.
Anomalies are important because they indicate significant but rare events that may have a detrimental impact on the system.
Therefore, they require prompt critical actions to be taken in a wide range of application domains.
An anomaly can be categorized in the following ways~\cite{ahmed2015investigation}.
\begin{itemize}
    \item Point anomaly: When a data instance deviates from the normal pattern of the dataset, it can be considered a point anomaly.
    \item Contextual anomaly: When a data instance behaves anomalously in a particular context, it is called a contextual or conditional anomaly.
    \item Collective anomaly: When a collection of similar data instances behave anomalously compared to the entire dataset, the group of data instances is called a collective anomaly.
\end{itemize}
Different supervised, unsupervised, and semi-supervised approaches have been proposed for this purpose.
These techniques, including classification based network anomaly detection such as support vector machine~\cite{balabine2017method}, Bayesian network~\cite{kruegel2003bayesian}, neural network~\cite{poojitha2010intrusion}, and rule-based approaches~\cite{yang2013rule}.
Statistical anomaly detection techniques, including mixture model, signal processing technique~\cite{thottan2003anomaly}, and principal component analysis (PCA)~\cite{shyu2003novel}.
Other category includes information theory-based and clustering-based~\cite{ahmed2019data}.
The proposed summarization approach is a general approach used in two scenarios: i) as the preprocessing approach where results are used as the input for anomaly detection algorithm, and ii) as an anomaly detection technique in a supervised setting discussed in the next section.
\section{The Proposed Approach (INSIDENT)}
\label{proposed}
This section discusses our proposed methodology.
At first, we define the problem and then discuss our algorithm.
\subsection{Problem Definition}
In this paper, $x_i$ is a sample vector and $X = [x_1, x_2,..., x_N]$ is traffic data consists of $N$ sample where $x_i \in R^d$ which $d$ denotes the number of features.
$K$ is the number of clusters, and cluster centroids are denoted by $c$.
$x_=$ is the closest similar sample to $x$, and
$x_\neq$ is the closest different sample.
An example of network traffic data with few attributes is reported in Table~\ref{tab:sample}.
The goal is to find a cluster of similar samples and find representatives for each cluster as the summary $S$  where they keep the same distribution but less in size.

\begin{table}[t]
\centering
\caption{Example of network traffic samples.}
\label{tab:sample}
    \begin{tabular}{|l|c|c|c|c|}
        \hline
        Source IP & Source Port & Destination IP & Destination Port  & Protocol \\
        \hline
        192.168.5.10 & 1234 & 192.168.1.1 & 80 & TCP\\
        \hline
        192.168.5.12 & 4565 & 192.168.1.2 & 20 & TCP\\
        \hline
        192.168.5.10 & 20 & 192.168.28.80 & 119 & HTTP\\
        \hline
        192.168.5.10 & 70 & 192.168.1.1 & 50 & TCP\\
        \hline
        211.204.12.10 & 31 & 192.168.28.80 & 119 & HTTP\\
        \hline
        192.168.5.1 & 3214  & 192.168.1.2 & 86 & TCP\\
        \hline
    \end{tabular}
\end{table}
\subsection{Methodology}
Previous approaches used different clustering or sampling algorithms to summarize data. 
However, there is no guarantee that the summarized data has the same distribution as the original data, and therefore as the substitute for the original data.
In this paper, we investigate the adaptation of clustering and the KNN algorithm to understand the data's underlying structure. 
In our previous work, this structure was used in the context of multi-document summarization~\cite{ghodratnama2020extractive} and image retrieval~\cite{ghodratnama2020content}, demonstrating promising results.
For this reason, the error rate of the nearest neighbor classifier in each cluster is minimized by locally weighting features in each cluster.
INSIDENT transforms the feature space into a new feature space by weighting features separately in each cluster, where outliers are recognized easier in the new feature space.
To this end, the weighted Euclidean distance is used.
In our problem, these weights are arranged in a $d\times K$ weight matrix $W=\{w_{ij}, 1\leq i\leq d, 1 \leq j \leq K\}$ where $d$ is the number of features, and is $K$ the number of clusters.
To be more specific, for each cluster we have a vector of weights corresponding to each feature which are representative of the importance of each feature in each cluster.
Our objective function is designed to minimize the error of 1NN in each cluster by regulating weights of each feature, and consequently cluster centers.
To estimate the error of 1NN the following approximation function defined in~\cite{paredes2006learning} is used:
\begin{equation}
J(\textbf{W})=\frac{1}{N}\sum_{s\in XS}^{}S_{\beta}(\frac{d_w(x,x_=)}{d_w(x,x_{\neq})})
\end{equation}
where the sample $x_=$ is the nearest similar sample, and the sample $x_{\neq}$ is the closest different sample to the input sample $x$. 
Respectively $d_w$ is the weighted Euclidean distance, and $S_{\beta}$ is the sigmoid function, defined as:
\begin{equation}
S_{\beta}(z)=(\frac{1}{1+{e}^{\beta(1-z)}})
\end{equation}
The objective function of K-means, which aims to minimize the errors of each cluster, is defined as:
\begin{equation}
J(\textbf{W},\textbf{C})=\sum_{k=1}^{K}\sum_{i=1}^{\mid N_K \mid} d_{W_{K}}^2 (x_i,c_K)
\end{equation}
Thus, the overall objective function is defined as:
\begin{dmath}
J(\textbf{W},\textbf{C})=  (\sum_{k=1}^{K}\sum_{i=1}^{\mid N_K \mid} d_{W_{}}^2 (s_i,c_K) + \frac{1}{N}\sum_{k=1}^{K}\sum_{i=1}^{\mid N_K \mid}S_{\beta}(\frac{d_w(x,x=)}{d_w(x,x_{\neq})}))
\end{dmath}
where the first term is the objective function of K-means, and the second term is the summation of the classification errors over the $K$ clusters. 

Two parameters are optimized in this objective function.
The first is the weights matrix. 
The feature-dependent weights associated with the sample $x_=$ are trained to make it closer to $x$, while making the sample $x_{\neq}$ further from $x$.
Then, the cluster centroid update is based on the learned weighted distance.
Since this function is differentiable, we can analytically use gradient descent for estimating the matrix $W$, guaranteeing convergence.
The iterative optimization of a learning parameter like w is given below.
\begin{equation}
W^{t+1}=W^{t}-\alpha(\frac{J(\textbf{W},\textbf{C})}{\delta(W)})
\end{equation}
To simplify the formula, the function $R(x)$ is defined \cite{paredes2006learning} as:
\begin{equation}
R_w(x_i)=(\frac{d_w(x_i,x_{i,=})}{d_w(x_i,x_{i,\neq})})
\end{equation}
The partial derivative of $J(W,C)$ with respect to $W$ is calculated by:
\begin{dmath}
    {\frac{\delta J(\textbf{W},\textbf{K})}{\delta W_K}} \cong \sum_{i=1}^{\mid N_K \mid} 2W_K \odot (x_i-C_K)^2+\frac{1}{N}\sum_{i=1}^{\mid N_K \mid} S_{\beta}^{'}(R(x_i))\frac{\delta R(x_i)}{\delta W_k}
\end{dmath}
where $\odot$ is the inner product and $\frac{\delta R(x_i)}{\delta W_K}$ is :
\begin{dmath}
    \frac{\delta R(s_i)}{\delta W_K}=\frac{1}{{}d_{W_K}^2}(x_i,x_{i,\neq})({\frac{1}{R(x_i)}W_K\odot (x_i-x_{i,=})^2}-R(x_i)W_K\odot{(x_i-x_{i,\neq})^2)}
\end{dmath}
The derivative of $S_{\beta}(z)$ is defined as:
\begin{dmath}
S_{\beta}(z){'}=\frac{\delta S_{\beta}(z)}{\delta z}=\frac{\beta e^{\beta(1-z)}}{(1+e^{\beta(1-z)})^2}
\end{dmath}
The partial derivative of $J(\textbf{W},\textbf{C})$ with respect to $C$ is calculated as:
\begin{dmath}
\frac{J(\textbf{W},\textbf{C})}{\delta C_k} \cong \sum_{i=1}^{\mid N_k \mid} -2W_k^2 \odot(x_i-C_k)
\end{dmath}


\begin{algorithm}[t]
\SetAlgoLined
 \SetKwInOut{Input}{input}
 \SetKwInOut{Output}{output}
 \Input{Traffic Data X, learning rate $\gamma$ and $\alpha$.}
 \Output{Summary ($S$).}\
 Ranked Sentences $\leftarrow$ ExDos(D)\;
    \While{$iter < Max Iterations$}{
    Clusters (C) $\leftarrow$ K-means(X)\;
     \For{each clusters c in C}{
        \For{each sample x  in c}{
            $x_{=} \leftarrow findSimilarCloseSample()$\;
            $x_{\neq} \leftarrow findDifferentCloseSample()$\;
             $\textit{$W^{iter+1}=W^{iter}-\gamma \frac{\delta J(W)}{W}$ }$\;
        }
     }
    Update Clusters\;
    }
return Summary(S)\;
\caption{INSIDENT}
\label{alg1}
\end{algorithm}

Since we need to optimize the weight of features for each cluster's samples, along with the center of clusters, we first update $ W $ in each cluster, and then we update $C$ (center of clusters).
The INSIDENT  algorithm is depicted in Algorithm~\ref{alg1} for more clarification.
Since the algorithm performs in an iterative process using gradient descent, the simplest clustering (k-means) and (KNN) algorithms are used for efficiency.
However, K-means is one of the most reliable and most widely used clustering algorithms.
Besides, the K-nearest neighbor (NN) has been successfully used in many pattern-recognition applications~\cite{anava2016k}. 
Similar samples are close to each other in new feature space, making a point, and contextual type anomalies easily detectable.
In the case of collective anomalies, we select the number of each cluster's representative based on its size to keep the distribution the same as the original data. 
\section{Experiments and Evaluation}
\label{evaluation}
In this section, the dataset, the evaluation method, and the performance of INSIDENT are explained and compared with existing state-of-the-art approaches.
\subsection{Data Set}
Experiments on six benchmark datasets are performed. 
The details of this dataset and the distribution of normal and anomalous samples in each dataset are reported in Table~\ref{tab:dataset}.
KDD1999 contains collective anomalies were the other five datasets contain only rare anomalies.
These rare anomalous datasets are from SCADA network, including real SCADA (WTP), simulated anomalies (Sim1 and Sim2), and injected anomalies (MI and MO).

\subsection{Evaluation Metrics}
To evaluate network traffic summary, we explain two widely used summary evaluation metrics including \textit{conciseness}, and \textit{information loss}~\cite{ahmed2015efficient}.
\begin{itemize}
    \item Conciseness: The size of the summary influences the quality of the summary.
    At the same time, it is important to create a summary that can reflect the underlying data patterns.
    Conciseness is defined as the ratio of input dataset size ($N$) and the summarized dataset size ($S$) defined as:
    \begin{equation}
        Conciseness=\frac{N}{S}
    \end{equation}
    \item Information Loss: A general metric used to describe the amount of information lost from the original dataset due to the summarization.
    Loss is defined as the ratio of the number of samples not present by samples present in summary defines as:
    \begin{equation}
        Information Loss=\frac{L}{T}
    \end{equation}
    where $T$ is the number of unique samples represented by the summary, and $L$ defines the number of samples not present in the summary.
\end{itemize}
\begin{table}[t]
\centering
\caption{Dataset Description.}
\label{tab:dataset}
    \begin{tabular}{|l|c|c|c|}
        \hline
        Dataset & Sample Number &  Normal Percentage & Anomalies Percentage\\
        \hline
        KDD1999 &494020 &19.69 &80.310\\
        \hline
        WTP &527 &97.34 &2.66\\
        \hline
        MI &4690 &97.86 &2.14\\
        \hline
        MO &4690 &98.76 &1.24\\
        \hline
        Sim1 &10501 &99.02 &0.98\\
        \hline
        Sim2 &10501 &99.04 &0.96\\
        \hline
    \end{tabular}
\end{table}
Besides, to evaluate the performance of the anomaly detection algorithms used in supervised approaches, three measures, including accuracy, recall, and F1 discussed below, are used.
Before we define these measure, four values included in the confusion needs to be discussed~\cite{ahmed2015investigation}.
\begin{itemize}
    \item True Positive (TP): Number of anomalies correctly identified as anomalous.
    \item False Positive (FP): Number of normal data incorrectly identified anomaly.
    \item True Negative (TN): Number of normal data correctly identified as normal.
    \item False Negative (FN): Number of anomalies incorrectly identified as normal.
\end{itemize}
Based on the above definitions, we define the evaluation metrics.

\begin{equation}
    Accuracy=\frac{TP+TN}{TP+TN+FP+FN}
\end{equation}

\begin{equation}
    Recall=\frac{TP}{TP+FN}
\end{equation}

\begin{equation}
    F1=\frac{2TP}{2TP+FP+FN}
\end{equation}

\subsection{Result Analysis}
In this section, we discuss the performance evaluation of the existing summarization methods compared to INSIDENT, along with the anomaly detection result.

\begin{table}[t]
\centering
\caption{Real SCADA dataset (WTP) result.}
\label{tab:WTP}
    \begin{tabular}{|l|c|c|c|}
        \hline
        Model & WTP-Recall &  WTP-Accuracy & WTP-F1\\
        \hline
        KNN & 85.71 &  97.39 & 85.71\\
        \hline
        LOF & 78.57 &  97.38 & 78.57\\
        \hline
        COF & 57.14 &  97.35 & 57.14\\
        \hline
        LOCI & 85.71 &  97.39 & 85.71\\
        \hline
        LoOP & 42.85 &  97.33 & 42.85\\
        \hline
        INFLO & 57.14 &  97.35 & 57.14\\
        \hline
         CBLOF & 92.85 &  97.40 & 92.85\\
        \hline
         LDCOF & 85.71 &  97.39 & 85.71\\
        \hline
         CMGOS & 57.14 &  97.35 & 57.14\\
        \hline
         HBOS & 28.57 &  97.32 & 28.57\\
        \hline
         LIBSVM & 85.71 &  97.39 &  85.71\\
        \hline
        \textbf{INSIDENT}&94.87 &97.91 &94.87\\
        \hline
    \end{tabular}
\end{table}

\subsubsection{Anomaly Detection Evaluation}
This section contains the performance analysis of anomaly detection techniques.
The baseline algorithms include Nearest Neighbor-based algorithms( K-NN~\cite{ramaswamy2000efficient}, LOF~\cite{breunig2000lof}, COF~\cite{tang2007capabilities}, LOCI~\cite{papadimitriou2003loci}, LoOP~\cite{kriegel2009loop}, INFLO~\cite{jin2006ranking}), clustering-based approach(CBLOF~\cite{he2003discovering}, LDCOF~\cite{amer2012nearest}, CMGOS~\cite{amer2012nearest}),  and statistical appraoches (HBOS and LIBSVM~\cite{amer2013enhancing}).
These approaches are compared with INSIDENT on different variations of the SCADA dataset, including WTP, MI, MO, Sim1, and Sim2, where their values are reported by~\cite{ahmed2015investigation}.
Results are reported respectively in Table~\ref{tab:WTP}, Table~\ref{tab:SIM}, and Table~\ref{tab:MI}.

From Table~\ref{tab:WTP}, it can be seen that for the real SCADA dataset(WTP), INSIDENT has higher values.
Then the clustering-based anomaly detection technique, CBLOF, performs best, and third, the nearest-neighbor-based approach attains the best performance.
It is an expected result showing the combination of clustering and KNN can perform better.
Statistical based approach HBOS does not perform well.
Table~\ref{tab:SIM} displays the results on simulated datasets (Sim1 and Sim2).
LIBSVM has better recall than others, and INCIDENT performs as the second best.
Clustering-based approaches are not well suited for the simulated datasets.
For the datasets with injected anomalies (MI, MO), INCIDENT, along with clustering-based approaches, are the best considering the evaluation measures.
Nearest neighbor-based approaches are the next best.
It is interesting to observe that the Recall and F1 values are identical for all the anomaly detection techniques.
The reason is that since the top $N$ anomalies detected by the techniques match the actual $N$ number of anomalies in the dataset, the Recall, and F1 scores are always the same.

\begin{table}[t]
\centering
\caption{Simulated SCADA datasets result(Sim1 and Sim2).}
\label{tab:SIM}
    \begin{tabular}{|l|c|c|c|c|c|c|}
        \hline
        Model & Sim1Recall &  Sim1Accuracy & Sim1F1 & Sim2Recall &  Sim2Accuracy & Sim2F1\\
        \hline
        KNN & 64.7&  99.03& 64.7 &63 &99.05&63\\
        \hline
        LOF & 0&  99.01 &0& 0 &99.03&0\\
        \hline
        COF & 0 &  99.01 & 0& 2&99.03&2\\
        \hline
        LOCI &0 &  99.01 & 0 & 0&99.03&0\\
        \hline
        LoOP & 0.98 &  99.01 & 0.98& 0&99.03&0\\
        \hline
        INFLO & 0 &  99.01 & 0& 0&99.03&0\\
        \hline
         CBLOF & 0 &  99.01 & 0& 0&99.03&0\\
        \hline
         LDCOF & 0 &  99.01 & 0& 0 &99.03&0\\
        \hline
         CMGOS & 18.62 & 99.02 & 18.62& 97&99.05&97\\
        \hline
         HBOS & 30.39 &  99.02 & 30.39 & 27&99.04&6\\
        \hline
         LIBSVM & 74.50 &  99.03 &  74.50 & 68&99.05&68\\
        \hline
        \textbf{INSIDENT} &72.13 &99.07 &72.13 &78.21 &99.05 &78.21\\
        \hline
    \end{tabular}
\end{table}
\vspace{-4mm}
\begin{table}[t]
\centering
\caption{Simulated SCADA datasets with Injected Anomalies result (MI and MO).}
\label{tab:MI}
    \begin{tabular}{|l|c|c|c|c|c|c|}
        \hline
        Model & MI-Recall &  MI-Accuracy & MI-F1 &  MO-Recall &   MO-Accuracy &  MO-F1\\
        \hline
        KNN &96 &97.09 &96 &91.37 &98.77 &91.37\\
        \hline
        LOF &38.33 &97.43 &38.33 &55.17 &98.76 &55.17\\
        \hline
        COF &9 &97.82 &9 &25.86 &98.75 &25.86\\
        \hline
        LOCI &91 &97.9 &91 &84.48 &98.77 &84.48\\
        \hline
        LoOP &10 &97.83 &10 &27.58 &98.75 &27.58\\
        \hline
        INFLO &12 &97.83 &12 &43.1 0&98.76 &43.10\\
        \hline
         CBLOF &24 &97.84 &24 &63.79 &98.76 &63.79\\
        \hline
         LDCOF &100 &97.91 &100 &63.79 &98.76 &63.79\\
        \hline
         CMGOS &100 &97.91 &100 &50 &98.76 &50\\
        \hline
         HBOS &98 &97.91 &98 &65.51 &98.76 &65.51\\
        \hline
         LIBSVM &86 &97.9 &86 &91.37 &98.77 &91.37\\
        \hline
        \textbf{INSIDENT} &100 &98.76 &100 &94.21 &99.04 &94.21\\
        \hline
    \end{tabular}
\end{table}
\subsubsection{Network Traffic Summarization Evaluation}
\begin{figure*}[t]
    \centerline{\includegraphics[width=0.65\textwidth]{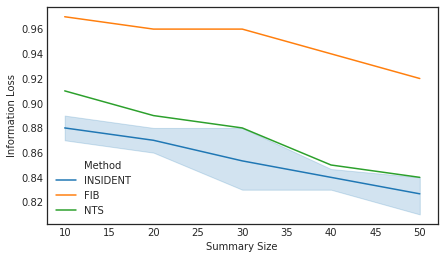}}
    \caption[]{The result of comparing information loss based on different summary size.}
    \label{fig:infoloss}
\end{figure*}

For summarization evaluation, the KDD dataset is used.
Summarization size, which defines conciseness, is considered as a constraint in summarization algorithms.
When the summary is small, it has maximum information loss.
On the other hand, when conciseness is small, the summary contains the whole dataset has no information loss. 
Therefore, information loss and conciseness are orthogonal parameters.
Our experiments used five different summary sizes, and then information loss was measured for each summary size.
In practice, the network manager/analyst decides the summary size based on the network.
The results are compared with NTS and FIB approaches~\cite{ahmed2014novel}.
Since our algorithm is based on k-means, we test three times with different initial points for each summary size.
Results are depicted in Figure~\ref{fig:infoloss}.
Besides, the percentage of anomalies compared with SUCh~\cite{ahmed2019intelligent} is reported in Table~\ref{tab:summary} proving that INSIDENT well-preserved the percentage of anomalies in generated summaries. 

\begin{table}[t]
\centering
\caption{Comparing the distribution of anomalies in summaries and original data.}
\label{tab:summary}
    \begin{tabular}{|l|c|c|c|}
        \hline
        Dataset &Original data &SUCh Alg. &INSIDENT\\
        \hline
        WTP &2.66 &N/A & 2.33\\
        \hline
        MI &2.14 &2.61 &2.16\\
        \hline
        MO &1.24 &1.46 &1.21\\
        \hline
        Sim1 &0.98 &1.04 &1.01\\
        \hline
        Sim2 &0.96 &0.94 &0.97\\
        \hline
    \end{tabular}
\end{table}
\section{Conclusion and Future Work}
\label{conclusion}
Monitoring network traffic data to detect any hidden patterns of anomalies is a challenging and time-consuming task which requires high computing resources.
Therefore, in this paper, we proposed an INtelligent Summarization approach for IDENTifying hidden anomalies, called \textit{INSIDENT}.
In data summarization, it is always a dilemma to claim the best summary.
The proposed approach claim is to guarantee to keep the original data distribution in summarized data. 
The INSIDENT's backbone is the clustering and KNN algorithm that dynamically maps original feature space to a new feature space by locally weighting features in each cluster.
The experimental results proved that the proposed approach helps keep the distribution the same as the original data, consequently making anomaly detection easier.
In future work, we aim to focus on real-time network traffic summarization.


\textbf{\emph{Acknowledgement.}}
We acknowledge the AI-enabled Processes (AIP) Research Centre  \footnote{https://aip-research-center.github.io/} for funding this research.
We also acknowledge Macquarie University for funding this project through IMQRES scholarship.
\bibliographystyle{splncs04}
\bibliography{Main.bib}
\end{document}